\documentclass[letterpaper]{article} % DO NOT CHANGE THIS
\usepackage{aaai2026}  % DO NOT CHANGE THIS
\usepackage{times}  % DO NOT CHANGE THIS
\usepackage{helvet}  % DO NOT CHANGE THIS
\usepackage{courier}  % DO NOT CHANGE THIS
\usepackage[hyphens]{url}  % DO NOT CHANGE THIS
\usepackage{graphicx} % DO NOT CHANGE THIS
\usepackage{natbib}  % DO NOT CHANGE THIS AND DO NOT ADD ANY OPTIONS TO IT
\usepackage{caption} % DO NOT CHANGE THIS AND DO NOT ADD ANY OPTIONS TO IT
\usepackage{algorithm}
\usepackage{algorithmic}
\usepackage{amssymb}
\usepackage{amsmath}
\usepackage{amsthm}
\usepackage{booktabs}
\usepackage{enumitem}
\usepackage{color}
\usepackage{tabularx}
\usepackage{tcolorbox}
\tcbuselibrary{skins}
\usepackage{subcaption}
\usepackage{multirow}
\usepackage{arydshln}
\usepackage{times}
\usepackage{soul}
\usepackage{url}
\usepackage[utf8]{inputenc}
\usepackage{pifont}
\newcommand{\cmark}{\ding{51}}%
\newcommand{\xmark}{\ding{55}}%
\usepackage{tikz}
\usetikzlibrary{shapes.geometric}

\newcommand{\circled}[1]{%
    \tikz[baseline=(node.base)]{%
        \node[draw,circle,inner sep=0.5pt,outer sep=0pt] (node) {#1};%
    }%
}
\usepackage{newfloat}
\usepackage{listings}
\DeclareCaptionStyle{ruled}{labelfont=normalfont,labelsep=colon,strut=off} % DO NOT CHANGE THIS
\floatstyle{ruled}
\newfloat{listing}{tb}{lst}{}
\floatname{listing}{Listing}
\title{LLM-based Contrastive Self-Supervised AMR Learning with Masked Graph Autoencoders for Fake News Detection}
\author {
    Shubham Gupta\textsuperscript{\rm 1}\equalcontrib,
    Shraban Chatterjee\textsuperscript{\rm 1}\equalcontrib,
    Suman Kundu\textsuperscript{\rm 1}
}
\affiliations {
    \textsuperscript{\rm 1}Indian Institute of Technology Jodhpur\\
    Jodhpur, Rajasthan, India\\
    gupta.37@iitj.ac.in, chatterjee.2@iitj.ac.in, suman@iitj.ac.in
}

\usepackage{bibentry}
\begin{document}

\maketitle

\begin{abstract}
The proliferation of misinformation in the digital age has led to significant societal challenges. Existing approaches often struggle with capturing long-range dependencies, complex semantic relations, and the social dynamics influencing news dissemination. Furthermore, these methods require extensive labelled datasets, making their deployment resource-intensive. In this study, we propose a novel self-supervised misinformation detection framework that integrates both complex semantic relations using Abstract Meaning Representation (AMR) and news propagation dynamics. We introduce an LLM-based graph contrastive loss (LGCL) that utilizes negative anchor points generated by a Large Language Model (LLM) to enhance feature separability in a zero-shot manner. To incorporate social context, we employ a multi view graph masked autoencoder, which learns news propagation features from social context graph. By combining these semantic and propagation-based features, our approach effectively differentiates between fake and real news in a self-supervised manner. Extensive experiments demonstrate that our self-supervised framework achieves superior performance compared to other state-of-the-art methodologies, even with limited labelled datasets while improving generalizability.
\end{abstract}

\section{Introduction}

\begin{table}
    \centering
     \caption{Comparison of different methods based on their utilization of various graph-based learning components. The table evaluates whether each method incorporates an AMR (Abstract Meaning Representation) graph, a Social Context Graph (SCG), a Graph Masked Autoencoder with augmentations (GMA$^2$), a Graph Masked Autoencoder with multi-view remasking (GMA$^2$+R), and Unsupervised Feature Generation (U).}
    \resizebox{\linewidth}{!}{%
    \begin{tabular}{c|cc|cc|c} \hline 
         Method & AMR & SCG & GMA$^2$ & GMA$^2$+R & U \\ \hline 
         EA$^2$N & \cmark & \xmark  & \xmark & \xmark & \xmark \\ 
         GACL & \xmark & \cmark & \xmark & \xmark & \xmark \\ 
         (UMD)$^2$ & \xmark & \cmark & \xmark & \xmark & \cmark \\ 
         GTUT & \xmark & \cmark & \xmark & \xmark & \cmark \\  
         GAMC & \xmark & \cmark & \cmark & \xmark & \cmark \\ \hline
         Ours & \cmark & \cmark & \cmark & \cmark & \cmark \\ \hline
    \end{tabular}
    }
    \label{tab:intro}
\end{table}

The spread of misinformation has become a significant problem in the digital age. It can lead to social unrest, foster hatred, erode trust, and ultimately impede the overall progress and stability of the society \cite{dewatana2021effectiveness}. Hence, effectively detecting misinformation has become an essential challenge to solve.

The concept of the ``veracity problem on the web'' was first introduced by \cite{yin2008} by designing a solution called \textit{TruthFinder}. This method verified news content by cross-referencing it with information from reputable websites. Later, \cite{feng-etal-2012-syntactic} employed manually crafted textual features for detecting misinformation. However, manually crafted features are time-consuming to create and fail to capture the complex semantic relations present in the text. Subsequently, many researchers turned to more advanced techniques, utilizing RNN's, and Transformer-based \cite{long-etal-2017-fake,Liu_Wu_2018} models to address this issue. For example, RNNs are employed to capture local and temporal dependencies within text data \cite{ma2016,li2022} and BERT has been increasingly utilized to improve the comprehension of contextual relationships in news articles \cite{devlin-etal-2019-bert}. Key limitations of these approaches are their struggle to maintain longer text dependencies and they do not capture complex semantic relations, such as events, locations, and trigger words. \cite{ea2n2025} solves this problem but requires supervision. Additionally, these models often neglect the social context and dynamics that influence news propagation \cite{yuan2019}. Acknowledging this, researchers have introduced graph-based approaches that integrate social context into the detection process \cite{min2022,sun2022,li2024}. Despite their effectiveness, these methods rely heavily on large, labelled datasets for training. Collecting and annotating such extensive datasets is time-consuming and resource-intensive, limiting their practical implementation. To address this \cite{Yin_Zhu_Wu_Gao_Wang_2024} propose a model to generate unsupervised features from the social context graph but do not consider the semantic relationship within the text. \textcolor{black}{Therefore, we require a model that is capable of incorporating semantic text features, a social context propagation graph and also perform well with minimal labelled data as highlighted in Table \ref{tab:intro}.}

This paper proposes a novel self-supervised misinformation detection methodology that considers complex semantic relations among entities in the news and the propagation of the news as a social context graph. In order to identify the semantic relations, this method incorporates a self-supervised Abstract Meaning Representation (AMR) encoder using the proposed graph contrastive loss. This loss creates feature separation by sampling negative anchor points using LLM. The use of negative anchor points from LLM helps in increasing the separation between fake and real classes in the latent space. In order to integrate the social context and capture the propagation of the news, our methodology also integrates a multi-view Graph Masked Autoencoder that employs the context and content of the news propagation process as the self-supervised signal to enhance the final feature space. These features, even with limited labelled data, achieve performance comparable or better than supervised counterparts using a simple linear SVM layer. The key contributions of our research are as follows:

% Specifically, we first create an abstract meaning representation of the news articles that represent semantic concepts as a Directed Acyclic Graph (DAG). The nodes in the graph are concepts such as entities, events, and attributes, and edges represent the relationships between those concepts. In order to reinforce verifiability, we infuse external evidence into the entity relationships \cite{ea2n2025}. Next, the AMR graph is passed to a self-supervised Graph Transformer that creates a latent feature space from the intricate semantic relationships present in the text. We update the latent space using our proposed contrastive loss, where we select a negative anchor against every sample with the help of an LLM in a zero-shot manner. Now that we have embedded the semantic relations in the text, we design another self-supervised multi-view masked graph encoder to capture the propagation of news in a social context. This approach uses feature reconstruction for graph self-supervised learning (SSL) through three strategies: \circled{1} graph augmentation, \circled{2} multi-view random re-mask decoding, and \circled{3} latent representation prediction. The multi-view random re-mask decoding denoises the input features and makes them robust to graph perturbation (using graph augmentation), thus synthesizing propagation features (latent representation prediction). Once the model is learned, these combined semantic and propagation features are used to predict the veracity of the news. 

\begin{itemize}
    \item A novel self-supervised learning based on AMR and social context graph is introduced in order to validate the veracity of news articles, eliminating dependence on labelled data. 
     \item In order to segregate the feature space among real and fake classes, graph contrastive loss is proposed. An LLM-based negative sampler is designed to handle negatives in the loss. 
     \item To capture the social context and propagation feature of the news, we propose an augmentation-based multi-view masked graph autoencoder module.
    \item Comprehensive evaluation with SOTA methods, demonstrating its superior performance.
\end{itemize}

\section{Related Work}
In this section, we provide a concise overview of the approaches utilized for detecting misinformation. The relevant studies are categorized into two main components: misinformation detection and self-supervised graph learning methodologies. %A detailed explanation of the advancements in each category is presented in the following subsections.

\subsection{Misinformation Detection Methods}
Early research on misinformation detection focused on manually crafted linguistic features \cite{feng-etal-2012-syntactic,jing2016,long-etal-2017-fake}, requiring significant effort for evaluation. EANN \cite{wang2018} is proposed to effectively extract event-invariant features from multimedia content, thereby enhancing the detection of misinformation on newly arrived events. In this line of work, recently, FakeFlow \cite{ghanem2021fakeflow} classified news using lexical features and affective information. In a separate line of work, external knowledge was integrated to improve model performance. Different source of external knowledge was used. For example, Popat et al. \cite{popat2017} retrieved external articles to model interactions; KAN \cite{dun2021} and CompareNet \cite{hu-etal-2021-compare} leveraged Wikidata for domain expansion, while KGML \cite{yao2021} bridged meta-training and meta-testing using knowledge bases. \textcolor{black}{Further, researchers have developed graph-based methods that incorporate social context into the detection process, for example, authors of GTUT \cite{gang2020} construct a graph for initial fake news spreader identification, (UMD)$^2$ \cite{silva2024} considers user credibility and propagation speed, GACL \cite{sun2022} constructs a tree of tweets for contrastive learning. All these methods do not leverage the complete propagation graph, and GACL requires supervision.  Other graph-based methods like \cite{min2022,li2024} rely heavily on manual annotation and external data. }

Recently, Abstract Meaning Representation (AMR)-based methods emerged to mitigate long-text dependency. Abstract Meaning Representation (AMR), as introduced by \cite{banarescu2013}, captures relationships between nodes using PropBank framesets. Recently, Zhang et al. \cite{zhang2023detecting} utilized AMR to detect out-of-context multimodal misinformation by identifying discrepancies between textual and visual data. In \cite{Gupta2023}, authors encoded textual information using AMR and explored how its semantic relations influence the veracity assessment of news. However, this study lacked sufficient evidence or justification for entity relationships within the AMR graph. Further, in the integration of evidence in AMR, EA$^2$N \cite{ea2n2025} is proposed that effectively captures evidence among entities present in AMR. All of these approaches rely on supervised data for AMR training and have not explored the potential of unsupervised methods.

%However, despite its extensive use in NLP, AMR has been underexplored for modeling complex semantic relationships between entities and providing evidence to support claims made in documents, especially for misinformation detection. AMR has been effectively applied in various NLP domains such as summarization \cite{kouris2022}, event detection \cite{wang-etal-2021-cleve}, and question answering \cite{lim-etal-2020-know} among others. , sentence vocabularies, and a wide range of over a hundred semantic relations, including negation, conjunction, command, and wikification. Its goal is to represent sentences with identical semantic meaning using the same AMR graph.
\subsection{Self-Supervised Graph Learning}
Self-supervised graph learning harnesses the structured richness of graph data to derive meaningful representations without relying on explicit labels \cite{wu2021}. A Graph Auto-Encoder (GAE) based model proposed that learns low-dimensional graph representations \cite{kipf2016variationalgraphautoencoders}. Later studies improved GAEs by focusing on reconstructing masked node features to enhance self-supervised learning for classification \cite{10.1145/3534678.3539321}. Further, \cite{10.1145/3543507.3583379} improved the performance by introducing multi-view random remasking. Recently, an unsupervised method for detecting misinformation GAMC \cite{Yin_Zhu_Wu_Gao_Wang_2024} has been proposed by leveraging both the context and content of news propagation as self-supervised signals. However, GAMC does not effectively handle complex semantic relations for longer text dependencies.

%Building upon the concept of a graph mask autoencoder, our work introduces a self-supervised graph autoencoder designed to generate representations of news, specifically tailored for the task of detecting misinformation.  Kipf et al. introduced a Graph Auto-Encoder (GAE), a method that encodes a graph into a lower-dimensional space and reconstructs it back to its original form, surpassing traditional approaches based on manually crafted features \cite{kipf2016variationalgraphautoencoders}. Recognizing that many GAEs struggle to reconstruct node features, subsequent research has focused on reconstructing masked features to improve the efficiency of self-supervised GAEs for classification tasks \cite{10.1145/3534678.3539321}.
\begin{figure*}
    \centering
    \includegraphics[width=.8\linewidth]{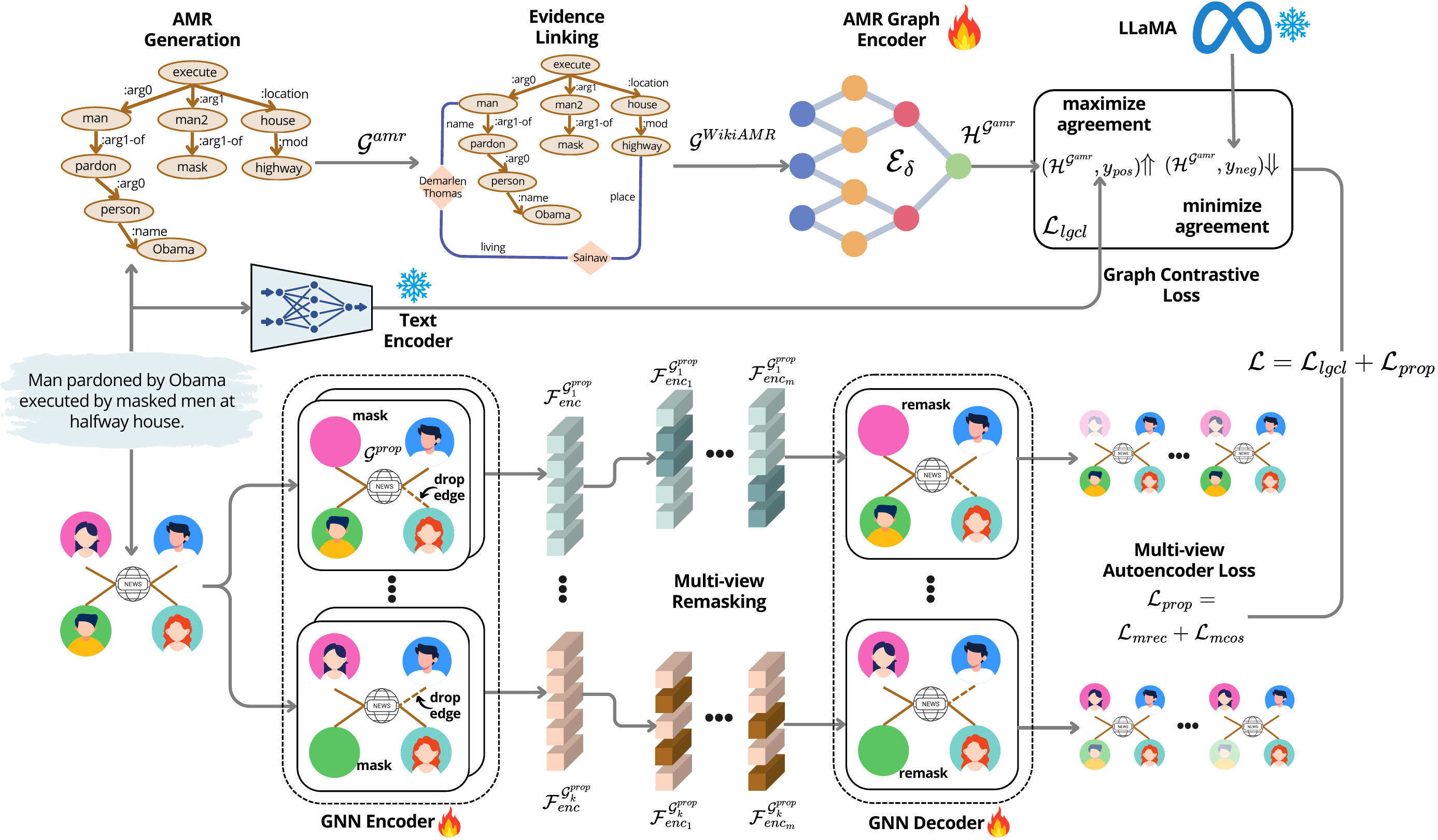}
    \caption{Overview of the proposed method: The news article is converted to an AMR graph $\mathcal{G}^{amr}$. $\mathcal{G}^{amr}$ is then linked to external evidences from Wikipedia represented as $\mathcal{G}^{WikiAMR}$. This $\mathcal{G}^{WikiAMR}$ graph is then converted to latent space features $\mathcal{H}^{\mathcal{G}^{amr}}$ by the graph transformer $\mathcal{E}_{\delta}$ based on $\mathcal{L}_{lgcl}$ optimization. The propagation graph of the same news article is then extracted and multiple augmentations are created. These augmented graphs are then passed to our multi-view remasked graph autoencoder which is optimized using $\mathcal{L}_{prop}$. The propagation graph feature $\mathcal{H}^{\mathcal{G}^{prop}}$ for each news is extracted from the trained GNN encoder. The final features for misinformation classification are obtained by concatenating $\mathcal{H}^{\mathcal{G}^{amr}}$ and $\mathcal{H}^{\mathcal{G}^{prop}}$.}
    \label{fig:model}
\end{figure*}
\section{Methodology}

The overall methodology is presented in Figure \ref{fig:model}. In this section we present these in more detail. 

\subsection{Self-supervised AMR Graph Learning}

%\subsubsection{AMR Parsing}
Given an input text $T$, we first create the AMR graph  $\mathcal{G}^{amr} (\mathcal{V}^{amr}, \mathcal{E}^{amr})$ capturing the relationships between different entities. AMR generation process involves parsing the sentences to extract linguistic information, including semantic roles, relations, and core events. In order to incorporate reasoning through AMR, we have integrated the external evidence by using the Evidence Linking Algorithm (ELA) used in \cite{ea2n2025}. The graph after applying ELA is referred to as WikiAMR, represented as $\mathcal{G}^{WikiAMR}$. In the paper, authors have shown the importance of WikiAMR over AMR. WikiAMR comprises interconnected undirected paths between entity nodes in  $\mathcal{G}^{amr}$ generated from the text. The WikiAMR representation helps to distinguish the difference between real and fake articles.

\paragraph{AMR Graph Learning with Path Optimization:}
This module plays an important role in extracting meaningful features from the given WikiAMR graph. Features extracted here capture essential semantic relationships, enabling a deeper understanding of the underlying textual data. At the core of this module is a Graph Transformer \cite{cai2020}, which employs various attention mechanisms to effectively process the graph representation. This allows the model to reason about and learn from the text more comprehensively.

The WikiAMR graph is first passed through a node initialization and relation encoder to transform it into a representation in $\mathbb{R}^{n\times k \times d}$,  where $n$, $k$, and $d$ denote the batch size, maximum sequence length, and the dimensionality of the graph encoding, respectively. 
To facilitate the model in identifying specific paths within $\mathcal{G}^{WikiAMR}$, the relation encoder computes the shortest path between two entities. This sequence of the path is subsequently converted into a relation vector using a Gated Recurrent Unit (GRU)-based RNN \cite{cho-etal-2014-learning}. $q_t$ is the sequence encoding extracted from GRU to get the relation vector $r_{uv}$. The mathematical formulation for this encoding is given by:

\begin{equation*}
\begin{aligned}
    \overrightarrow{q}_t &= \mathrm{GRU}_{f}(\overrightarrow{q}_{t-1}, sp_{t}) \\
    \overleftarrow{q}_t &= \mathrm{GRU}_{b}(\overleftarrow{q}_{t+1}, sp_{t})
\end{aligned}
\end{equation*}

Here, $sp_t$ represents the shortest path between two entities. Formally, the shortest relation path $sp_{i \rightarrow j}$ $ = [e(u, k_1), e(k_1, k_2), \ldots, e(k_n, v)]$ between the node $u$ and the node $v$, where $e(\cdot, \cdot)$ indicates the edge label and $k_{1:n}$ are the relay nodes. To compute the attention scores, the final relational encoding $r_{uv}$ is split into two distinct components, $r_{u \to v}$ and $r_{v \to u}$, via a linear transformation with a parameter matrix $W_r$:

\begin{equation*}
r_{uv} = [\overrightarrow{q}_n; \overleftarrow{q}_0], \quad 
[r_{u \to v}; r_{v \to u}] = W_r r_{uv}
\end{equation*}

Subsequently, attention scores $\beta_{uv}$ are calculated by incorporating both entity and relation representations from the graph $\mathcal{G}^{WikiAMR}$:

\begin{equation}
\begin{aligned}
&\beta_{uv} = h(e_u, e_v, r_{uv}) \\
&= \; (e_u + r_{u \to v}) W_p^\top W_k (e_v + r_{v \to u}) \\
&= \; \underbrace{e_u W_p^\top W_k e_v}_{\circled{a}} 
\; + \; \underbrace{e_u W_p^\top W_k r_{v \to u}}_{\circled{b}} \\
&\; + \; \underbrace{r_{u \to v} W_p^\top W_k e_v}_{\circled{c}} 
\; + \; \underbrace{r_{u \to v} W_p^\top W_k r_{v \to u}}_{\circled{d}}
\end{aligned}
\label{eq:attention_scores}
\end{equation}

The attention weights computed here guide the focus on entities according to their relationships. Each term in Equation \ref{eq:attention_scores} serves a distinct purpose: (a) models content-based attention, (b) captures biases related to the source of the relationship, (c) addresses biases from the target, and (d) encodes a general relational bias, providing a comprehensive view of entity interactions.
Finally, the Graph Transformer $(\mathcal{E}_\delta)$ encodes $\mathcal{G}^{WikiAMR}$, producing the final graph representation as follows:

\begin{equation}
\mathcal{H}^{\mathcal{G}^{amr}} = \mathcal{E}_\delta(\mathcal{G}^{WikiAMR}) \in \mathbb{R}^{n \times k \times d}
\end{equation}

Here, $\mathcal{H}^{\mathcal{G}^{amr}}$ represents the output graph embeddings generated by the Graph Transformer, and $d$ is the feature dimensionality.

\paragraph{Graph Contrastive Loss:}
Our proposed LLM-based graph contrastive loss (LGCL) function comprises two primary objectives. The first objective aims to ensure that the graph embedding remains close to its original embedding space by minimizing the reconstruction error between the predicted feature and the original feature. The second objective seeks to maximize the divergence between the predicted feature and the negative sample feature. To quantify the similarity between features, we utilize the Scaled Cosine Error (SCE) \cite{10.1145/3534678.3539321}. Formally, given the original feature $Y$ and the reconstructed output $Y'$, SCE is defined as:

\begin{equation}
\mathcal{L}_{\mathrm{SCE}} = \frac{1}{|\mathcal{N}|} \sum_{n \in \mathcal{N}} \left( 1 - \frac{y_i^T y'_i}{\|y_i\| \cdot \|y'_i\|} \right)^\gamma, \quad \gamma \geq 1
\end{equation}
Here, $\gamma$ is a scaling factor. When predictions have high confidence, the resulting cosine errors are generally less than 1 and diminish more quickly towards zero as the scaling factor $\gamma > 1$.

The contrastive loss requires both a positive sample feature $y_{\mathrm{pos}}$ and a negative sample feature $y_{\mathrm{neg}}$ to compare against the predicted feature. In the proposed formulation, $\mathcal{H}^{\mathcal{G}^{amr}}$ is used as $y'$, $y_{\mathrm{pos}}$ is the original BERT-derived feature of the input text, while $y_{\mathrm{neg}}$ is a negative sample feature generated using an LLM-based negative sampler.
The final contrastive loss for graph-based self-supervised learning (SSL) is formulated as follows:

\begin{equation}
\begin{split}
\label{eqn:loss2}
  \mathcal{L}_{lgcl} = &\mathcal{L}_{\mathrm{SCE}}(y', y_{\mathrm{pos}}) \\ &+ \lambda \cdot \max\left(0, m - \mathcal{L}_{\mathrm{SCE}}(y', y_{\mathrm{neg}})\right)  
\end{split}
\end{equation}
Here, $\lambda$ is a weighting factor, and $m$ is the margin to ensure negatives are pushed apart in cosine space.

% \subsection{LLM based Negative Sampler}
% In order to select the negative sample $y_{neg}$ for the contrastive learning, we have used Llama3-7B, a large language model, to reason over the given input samples. We have queried the Llama for assigning the fake and real labels to the input samples. Based on LLM's label, we find the centroid of the fake and real samples in order to use as a negative sample in the final loss.
\paragraph{LLM-based Negative Sampler:}
 We employ a large language model (LLM) in zero-shot to facilitate effective contrastive learning. Specifically, LLaMA3-7B is used to generate negative samples (\(y_{\text{neg}}\)). \textcolor{black}{This approach leverages the reasoning capabilities of the LLM to distinguish between real and fake input samples, assigning them pseudo labels for the selection of the negative feature for the contrastive learning task.}

Let \(\mathcal{X} = \{x_1, x_2, \ldots, x_n\}\) denote the set of input features. The input prompt and output format used for the LLM is mentioned in the end of the section. For each input \(x_i \in \mathcal{X}\), the LLM assigns a pseudo label \(\widetilde{y}_i \in \{0, 1\}\), where:
\[
\widetilde{y}_i =
\begin{cases} 
1 & \text{if } x_i \text{ is labelled as real}, \\
0 & \text{if } x_i \text{ is labelled as fake}.
\end{cases}
\]

Using the LLM's output labels, we partition the input samples into two groups:
\[
\mathcal{X}_{\text{real}} = \{x_i \mid \widetilde{y}_i = 1\}, \quad \mathcal{X}_{\text{fake}} = \{x_i \mid \widetilde{y}_i = 0\}.
\]

We compute the centroids of the real and fake samples as, 

\[
c_{\text{real}} = \frac{1}{|\mathcal{X}_{\text{real}}|} \sum_{x_i \in \mathcal{X}_{\text{real}}} \mathbf{f}_i, \quad
c_{\text{fake}} = \frac{1}{|\mathcal{X}_{\text{fake}}|} \sum_{x_i \in \mathcal{X}_{\text{fake}}} \mathbf{f}_i.
\]
where a feature vector \(\mathbf{f}_i \in \mathbb{R}^{n \times k \times d}\) is the initial BERT feature corresponding to \(x_i\). 
The negative sample (\(y_{\text{neg}}\)) is chosen to maximize the contrastive loss. In particular, we use \(c_{\text{fake}}\) as the representative negative sample for the real input sample, while \(c_{\text{real}}\) is used as the negative sample for the fake input sample. By leveraging the LLM to reason over input samples and compute these centroids, our approach effectively selects meaningful negative samples, enhancing the discriminative power of the contrastive learning model.

\begin{tcolorbox}[ width=\columnwidth,
                  interior hidden,
                  boxsep=3pt,
                  left=5pt,
                  right=5pt,
                  top=5pt,
                  bottom=1pt]
\textbf{LLM's Zero Shot Input Prompt:}\\
 Write in one word among `real' or `fake' whether given text is real or fake. \{text\}
 \\ \\
 \textbf{LLM's Output:} fake/real
\end{tcolorbox}

\subsection{Multi-View Social Context and Propagation Graph Learning}
%\subsection{Social Context Graph Propagation Learning}
% We introduce the self-supervised method that tries to learn the latent feature representation from a news propagation graph.  We denote the propagation graph for each news as $P_i$ where $P_i \in \{P_1, ..., P_n\}$. In this Section we learn a self-supervised encoder $\mathcal{E}$ that map each news node to a set $\{F, R\}$ where $F$ represents $F$ and $R$ represents real i.e. $\mathcal{E}: P_i \rightarrow \{F, R\}$.

%\paragraph{Graph Creation}
Each news article is converted into a propagation graph $G^{prop} = (V,E,\mathcal{F})$ as in \cite{10.1145/3404835.3462990}. Nodes in $V$ represent one news article and users who forward that article. An edge in $E$ exists between two nodes if there exists a forwarding relationship between them. The features for the news node are generated by passing the news article to a pre-trained language model (BERT), and the features for the user nodes are generated based on their recent 200 posts. The news and user node features are collectively referred to as $\mathcal{F}$. 

\paragraph{Graph Augmentation:}
We use two augmentation strategies: \circled{1} feature masking and \circled{2} random edge removal for creating augmentations of the input graph as suggested in \cite{Yin_Zhu_Wu_Gao_Wang_2024}. For input feature masking, we randomly select $50\%$ nodes in the graph and replace their features with a masked token. For \circled{2}, we randomly remove $20 \%$ edges from the graph. Each augmented graph for $G^{prop}$ is denoted as $\mathcal{G}^{prop}_{i}$. 
% Graph augmentation-based self-supervised learning enhances the robustness and generalization of graph neural networks by leveraging intrinsic structural and semantic properties, enabling models to learn richer representations without reliance on labelled data, \note{as demonstrated in}{This part is also taken from this paper? Or is there anything we contributed?} \cite{10.1145/3543507.3583379}.

\paragraph{Graph Encoding:}
We encode each $\mathcal{G}^{prop}_{i}$ into a latent space representation using a GNN encoder. For this, we use GIN \cite{xu2018how} represented using Equation \ref{eqn:GIN} as it is theoretically proven to distinguish between graph structures. 
\begin{equation}
    \label{eqn:GIN}
    f_v^{(k)} = \text{MLP} \left( (1 + \epsilon) \cdot f_v^{(k-1)} + \sum_{u \in \mathcal{N}(v)} f_u^{(k-1)} \right)
\end{equation}
Here, \( f_v^{(k)} \) is embedding of node \( v \) at layer \( k \), \( \mathcal{N}(v) \) contains neighbors of node \( v \) and \( \epsilon \) is a learnable scalar controlling residual connections. The final node embeddings from the encoder for each ${\mathcal{G}^{prop}_{i}}$ is represented as $\mathcal{F}_{enc}^{\mathcal{G}^{prop}_{i}}$. 

For downstream classification tasks on $G^{prop}$ we use the graph embedding \( \mathcal{H}^{{G}^{prop}} \) calculated as:
\begin{equation}
    \mathcal{H}^{{G}^{prop}} = \frac{1}{|V|} \sum_{v \in V} {f}_{v} \in \mathcal{F}_{enc}^{{G}^{prop}}
\end{equation}
% where \( V \) represents the set of nodes in the graph and \( h_v^{(K)} \) is the final-layer embedding of node \( v \). This averaging operation ensures that the graph representation captures the overall structural and feature-based characteristics of the individual nodes while maintaining permutation invariance. By aggregating the node representations in this manner, we generate a informative latent representation of the entire graph, which can be used for misinformation classification.

% \subsection{Graph Decoding}
% Now, that we have constructed the graph features from the input graph we again mask the graph features multiple times using random remasking. Remasking in the decoder stage of helps improve feature reconstruction by forcing the model to learn robust and generalizable representations of the underlying graph structure, rather than memorizing specific node embeddings as shown in \cite{10.1145/3543507.3583379}. 
\paragraph{Multi-View Graph Decoding:}
Now, from the encoded node representations $\mathcal{F}_{enc}^{\mathcal{G}^{prop}_{i}}$, we decode the input node features $\mathcal{F}$ using GIN as a decoder. In \cite{Yin_Zhu_Wu_Gao_Wang_2024} the authors use a single stage remasking for each $\mathcal{F}_{enc}^{\mathcal{G}^{prop}_{i}}$ to reconstruct the input features. But authors in \cite{10.1145/3543507.3583379} have shown that feature reconstruction is susceptible to congruence among the input features, which single remasking cannot address. To address this, we introduce multi-view feature remasking of each augmented graph $\mathcal{F}_{enc}^{\mathcal{G}^{prop}_{i}}$. Each remasked encoded feature is denoted by $\mathcal{F}_{enc_{j}}^{\mathcal{G}^{prop}_{i}}$. It acts as a regularizer for the decoder, making it robust against unexpected noises in input and helping to avoid overfitting. The final objective of the decoder is to reconstruct the actual node features $\mathcal{F}$ from these masked encoded node features using the multi-view autoencoder loss described next.

\paragraph{Multi-View Autoencoder Loss:}
Given $k$ augmentations of the input graph $\mathcal{G}^{prop}$ represented as $\mathcal{G}^{prop}_{1},\dots,\mathcal{G}^{prop}_{k}$, and $m$ remasked decoded output for each augmented graph represented as $\mathcal{F}_{dec_{1}}^{\mathcal{G}^{prop}_{1}}, \dots,\mathcal{F}_{dec_{m}}^{\mathcal{G}^{prop}_{1}}, \dots, \mathcal{F}_{dec_{m}}^{\mathcal{G}^{prop}_{k}}$, we define the multi-view reconstruction loss as
\begin{equation}
    \label{eqn:rec_loss}
    \mathcal{L}_{mrec} = \sum_{i=1}^{k}\sum_{j=1}^{m}(\mathcal{F}-\mathcal{F}_{dec_{j}}^{\mathcal{G}^{prop}_{i}})
\end{equation}
To minimize the divergence across the views of the decoded features, we define the multi-view cosine similarity loss as
\begin{equation}
    \label{eqn:con_loss}
    % \underset{\underset{l<k}{if \text{ }l = l^{'}  \text{then }i \neq j }}{\mathcal{M}}
    \mathcal{L}_{mcos} = \underset{\substack{\forall l, i, j; \text{ if } l = l' \text{ then } i \neq j \\l\leq k,i\leq m,j\leq m}}{\mathcal{M}}\frac{\mathcal{F}_{dec_{i}}^{\mathcal{G}^{prop}_{l}}.\mathcal{F}_{dec_{j}}^{\mathcal{G}^{prop}_{l^{'}}}}{\Big|\Big|\mathcal{F}_{dec_{i}}^{\mathcal{G}^{prop}_{l}}\Big|\Big|.\Big|\Big|\mathcal{F}_{dec_{j}}^{\mathcal{G}^{prop}_{l^{'}}}\Big|\Big|}
\end{equation}
Here, $\mathcal{M}$ is the mean operation. Our final propagation loss is $\mathcal{L}_{prop} = \mathcal{L}_{mrec} + \mathcal{L}_{mcos}$.
\subsection{Final Loss} 
We combine the AMR and Propagation loss as $\mathcal{L} = \mathcal{L}_{lgcl} + \mathcal{L}_{prop}$. We train our model using this loss, and the final features of our model are $\mathcal{H}^{\mathcal{G}^{amr}}\cdot\mathcal{H}^{\mathcal{G}^{prop}}$. These features are then used for misinformation classification.
\section{Experiments and Results}
% Experiments have been performed to compare the results of the proposed method with different unsupervised and supervised baselines on two publicly available datasets. The details of the experiments and results are presented in this section. 

% \paragraph{Dataset Description:}
We perform experiments on the publicly available datasets FakeNewsNet \cite{kai2020} in order to assess the effectiveness of the model. This repository contains two separate benchmark datasets, namely, PolitiFact and GossipCop. Further details on the datasets and implementation of our model are provided in the supplementary document.

\paragraph{Baselines:}
In our evaluation, we contrast our model with various state-of-the-art baselines, categorized into two groups. The first group utilizes only unsupervised methods (\textbf{TruthFinder} \cite{yin2008}, \textbf{UFNDA} \cite{li2021}, \textbf{UFD} \cite{yang2022}, \textbf{GTUT} \cite{gang2020}, \textbf{(UMD)$^2$} \cite{silva2024}, \textbf{GraphMAE} \cite{10.1145/3534678.3539321}, \textbf{GAMC} \cite{Yin_Zhu_Wu_Gao_Wang_2024}), while the second incorporates supervised methods (\textbf{SAFE} \cite{zhou2020},  \textbf{EANN} \cite{wang2018}, \textbf{dEFEND} \cite{shu2019}, \textbf{GACL} \cite{sun2022}, \textbf{EA$^2$N (BERT)} \cite{ea2n2025}). 

\section{Results}
We conducted a comparative analysis of our model against various unsupervised and supervised baselines on the PolitiFact and GossipCop datasets. As shown in Table \ref{tab:unsupervised_pol}, our model achieved the highest accuracy ($0.919$), precision ($0.933$), recall ($0.903$), and F1-score ($0.918$) among the unsupervised baselines. Compared to GAMC, the existing benchmark, our model outperforms it by a margin of $8.1\%$ in accuracy and $8.7\%$ in F1-score (on the absolute scale). Also, our model surpasses GTUT and (UMD)$^2$ by significant margins, $12$ $\sim$ $14\%$ in accuracy and $14$ $\sim$ $15\%$ in the F1-score, indicating a superior ability to differentiate between fake and real news. In a similar context, as shown in Table \ref{tab:unsupervised_gos}, our model significantly outperforms existing unsupervised baselines on the GossipCop dataset. It achieves the highest accuracy ($0.968$), precision ($0.965$), recall ($0.967$), and F1-score ($0.966$), outperforming GAMC, which attained an accuracy of $0.946$ and an F1-score of $0.943$. This represents a $2.2\%$ improvement in accuracy and a $2.3\%$ improvement in the F1-score. This improvement can be attributed to the proposed model, which leverages a combination of self-supervised AMR semantic features and news propagation features from multi-view social context graph learning.

When we compare our model to supervised baselines on both PolitiFact and GossipCop datasets (Table \ref{tab:supervised}), it consistently outperforms state-of-the-art approaches in terms of accuracy, while comparable results on F1 score are observed. On PolitiFact, our model achieves an accuracy of $0.919$ and an F1-score of $0.933$, surpassing EA$^2$N with BERT ($0.911$ accuracy, $0.915$ F1-score), GACL ($0.867$ accuracy, $0.866$ F1-score), and EANN ($0.804$ accuracy, $0.798$ F1-score). However, it shows comparative performance with dEFEND in F1-score. On GossipCop, our model outperforms all supervised baselines, achieving the highest accuracy ($0.968$) and F1-score ($0.966$). It notably surpasses GACL ($0.907$ accuracy, $0.905$ F1-score) and EA$^2$N ($0.844$ accuracy, $0.872$ F1-score), as well as dEFEND, which lags significantly behind with $0.808$ accuracy and $0.755$ F1-score. These results highlight that while supervised models perform well, our self-supervised approach not only competes effectively on PolitiFact but outperforms all supervised baselines on GossipCop, demonstrating superior performance across datasets. Our self-supervised pipeline may yield stronger representations than shallow supervised models trained only on labels. One reason is that the datasets have known issues with label reliability. In such cases, supervised models can overfit to spurious correlations or unreliable labels and unsupervised models often rely on representation learning, which can be more robust to noise and generalize better in low-label regimes.

\label{sec:res}
\begin{table}[ht]
\centering
\caption{Comparative study of our model w.r.t. different unsupervised baselines on PolitiFact dataset.}
\label{tab:unsupervised_pol}
\resizebox{\linewidth}{!}{%
\begin{tabular}{c|cccc}
\hline  Methods & Acc & Pre & Rec & F1 \\
\hline TruthFinder & 0.581 & 0.572 & 0.576 & 0.573 \\
UFNDA & 0.685 & 0.667 & 0.659 & 0.670 \\
UFD & 0.697 & 0.652 & 0.641 & 0.647 \\
GTUT & 0.776 & 0.782 & 0.758 & 0.767 \\
(UMD)$^2$ & 0.802 & 0.795 & 0.748 & 0.761 \\
GraphMAE & 0.643 & 0.658 & 0.641 & 0.649\\
 GAMC & 0.838 & 0.836 & 0.827 & 0.831 \\
\hline Ours & \textbf{0.919}& \textbf{0.933}& \textbf{ 0.903}& \textbf{0.918}\\
variance & $\pm$ 0.019& $\pm$ 0.045& $\pm$ 0.058& $\pm$ 0.020\\
\hline 
\end{tabular}
}
\label{tab2}
\end{table}

\begin{table}[ht]
\centering
\caption{Comparative study of our model w.r.t. different unsupervised baselines on GossipCop dataset.}
\label{tab:unsupervised_gos}
\resizebox{\linewidth}{!}{%
\begin{tabular}{c|cccc}
\hline  Methods & Acc & Pre & Rec & F1 \\
\hline TruthFinder & 0.668 & 0.669 & 0.672 & 0.669 \\
UFNDA & 0.692 & 0.687 & 0.662 & 0.673 \\
UFD & 0.662 & 0.687 & 0.654 & 0.667 \\
GTUT & 0.771 & 0.770 & 0.731 & 0.744 \\
(UMD)$^2$ & 0.792 & 0.779 & 0.788 & 0.783 \\
GraphMAE & 0.802 & 0.781 & 0.793 & 0.787\\
 GAMC & 0.946 & 0.941 & 0.946 & 0.943 \\
\hline Ours & \textbf{0.968} & \textbf{0.965}& \textbf{0.967}& \textbf{0.966}\\
variance & $\pm$ 0.015& $\pm$ 0.026& $\pm$ 0.039& $\pm$ 0.015\\
\hline 
\end{tabular}
}
\end{table}

\begin{table}[ht]
\centering
\caption{Comparative study of our model with supervised methods on PolitiFact and GossipCop datasets.}
\label{tab:supervised}
\resizebox{.8\linewidth}{!}{%
\begin{tabular}{c|cc|cc}
\hline \multirow{2}{*}{ Dataset } & \multicolumn{2}{|c|}{ PolitiFact } & \multicolumn{2}{c}{ GossipCop } \\
\cline { 2 - 5 } & Acc & F1 & Acc & F1 \\
\hline SAFE & 0.793 & 0.775 & 0.832 & 0.811 \\
EANN & 0.804 & 0.798 & 0.836 & 0.813 \\
dEFEND & 0.904 &\textbf{0.928} & 0.808 & 0.755 \\
GACL & 0.867 & 0.866 & 0.907 & 0.905 \\
EA$^2$N & 0.911 & 0.915 & 0.844 & 0.872\\
\hline
Ours& \textbf{0.919} & 0.918 & \textbf{0.968}&\textbf{0.966}\\
\hline
\end{tabular}
}
\end{table}

\section{Ablation Study}
\begin{figure}
    \centering
    \includegraphics[width=.8\linewidth]{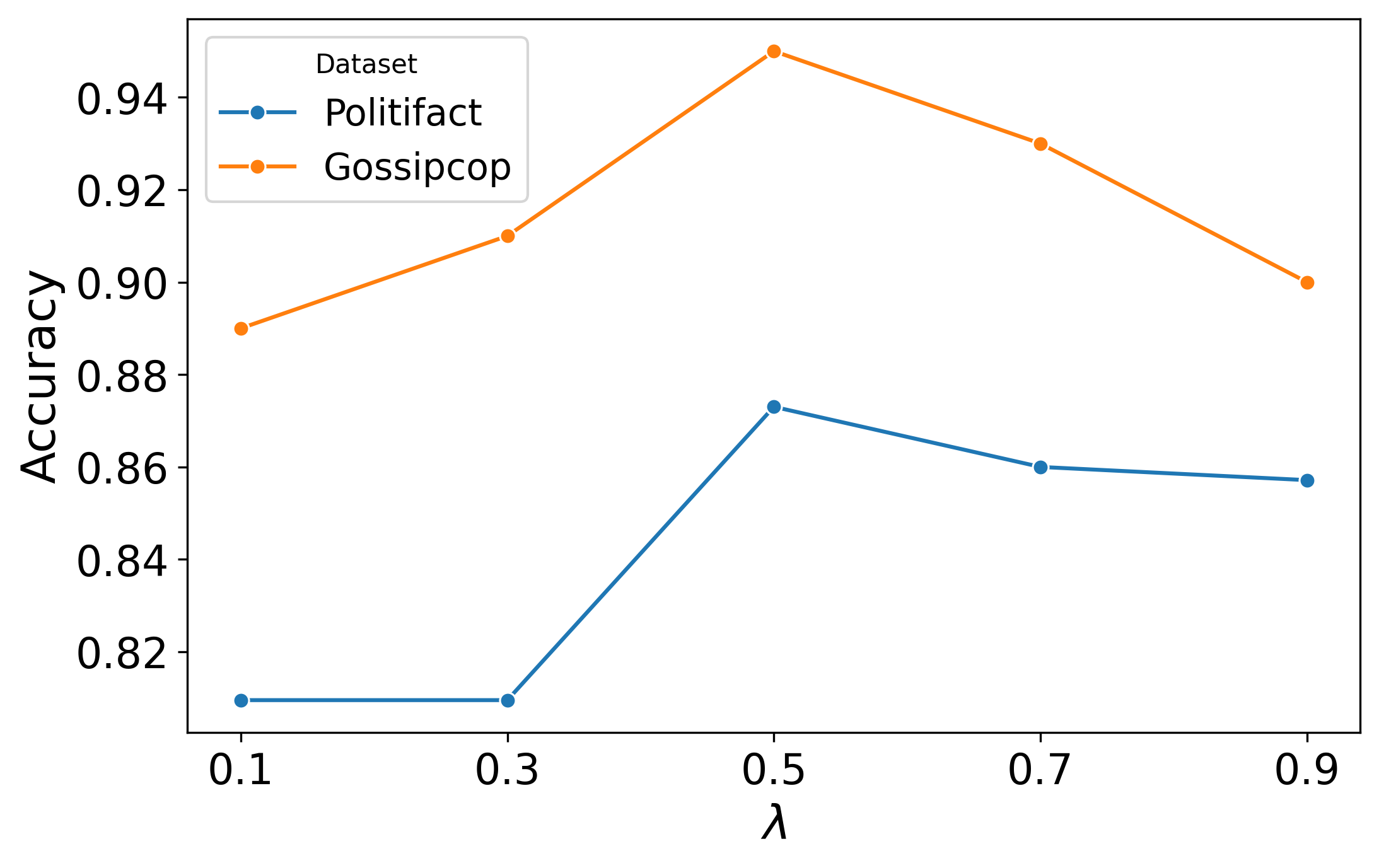}
    \caption{Change in classification result with different values of $\lambda$.}
    \label{fig:abl1}
\end{figure}

\paragraph{Change in classification result with different values of $\lambda$:} Figure \ref{fig:abl1} shows the change in classification accuracy of the method with the change in weightage to negative samples in Equation \ref{eqn:loss2}. It is evident that the accuracy improved initially with the value of $\lambda$ and obtained the maximum result when $\lambda = 0.5$ for both datasets. With a further increase in $\lambda$, the accuracy decreases, indicating that our model overemphasizes negative samples compared to being close to positive samples, thus decreasing feature separability. Based on this,  we set the value of $\lambda$ to $0.5$ in our experiments. 
\paragraph{Change in classification result with training size:} We study the effect of our features on misinformation We conduct a classification experiment using a linear SVM with varying training sizes while keeping the test set fixed at 10\%, as shown in Table \ref{abl:split_pol_gos}. The results clearly demonstrate that the classification accuracy improves with larger training data, which is consistent with expectations. Notably, our proposed model consistently outperforms traditional unsupervised baselines even with limited training samples, particularly in low-resource settings. With just 10\% of the training data, our model achieves superior performance on both the GossipCop and PolitiFact, highlighting its effectiveness in data-scarce scenarios. This showcases the robustness and generalization ability of the learned representations.

\begin{table}
    \centering
    \caption{Results on different split sizes for PolitiFact and GossipCop datasets.}
    \label{abl:split_pol_gos}
    \resizebox{0.9\linewidth}{!}{%
    \begin{tabular}{c|c|c|c|c} \hline 
         \multirow{2}{*}{Train Size \%}&  \multicolumn{2}{|c|}{PolitiFact}&  \multicolumn{2}{|c}{GossipCop}\\ 
         \cline { 2 - 5 } &  Acc&  F1&  Acc& F1\\ \hline 
         10&  0.875&  0.867&  0.951& 0.951\\ \hline 
         20&  0.875&  0.867&  0.948& 0.949\\ \hline 
         30&  0.875&  0.867&  0.951& 0.951\\ \hline 
         40&  0.906&  0.903&  0.952& 0.953\\ \hline 
         50&  0.906&  0.903&  0.952& 0.953\\ \hline 
         60&  0.906&  0.903&  0.952& 0.953\\ \hline 
         70&  0.906&  0.909&  0.952& 0.953\\ \hline 
         80&  0.938&  0.938&  0.954& 0.955\\ \hline 
         90&  0.938&  0.941&  0.956& 0.957\\ \hline
    \end{tabular}
    }
\end{table}

\begin{figure}
    \centering
    \includegraphics[width=.7\linewidth]{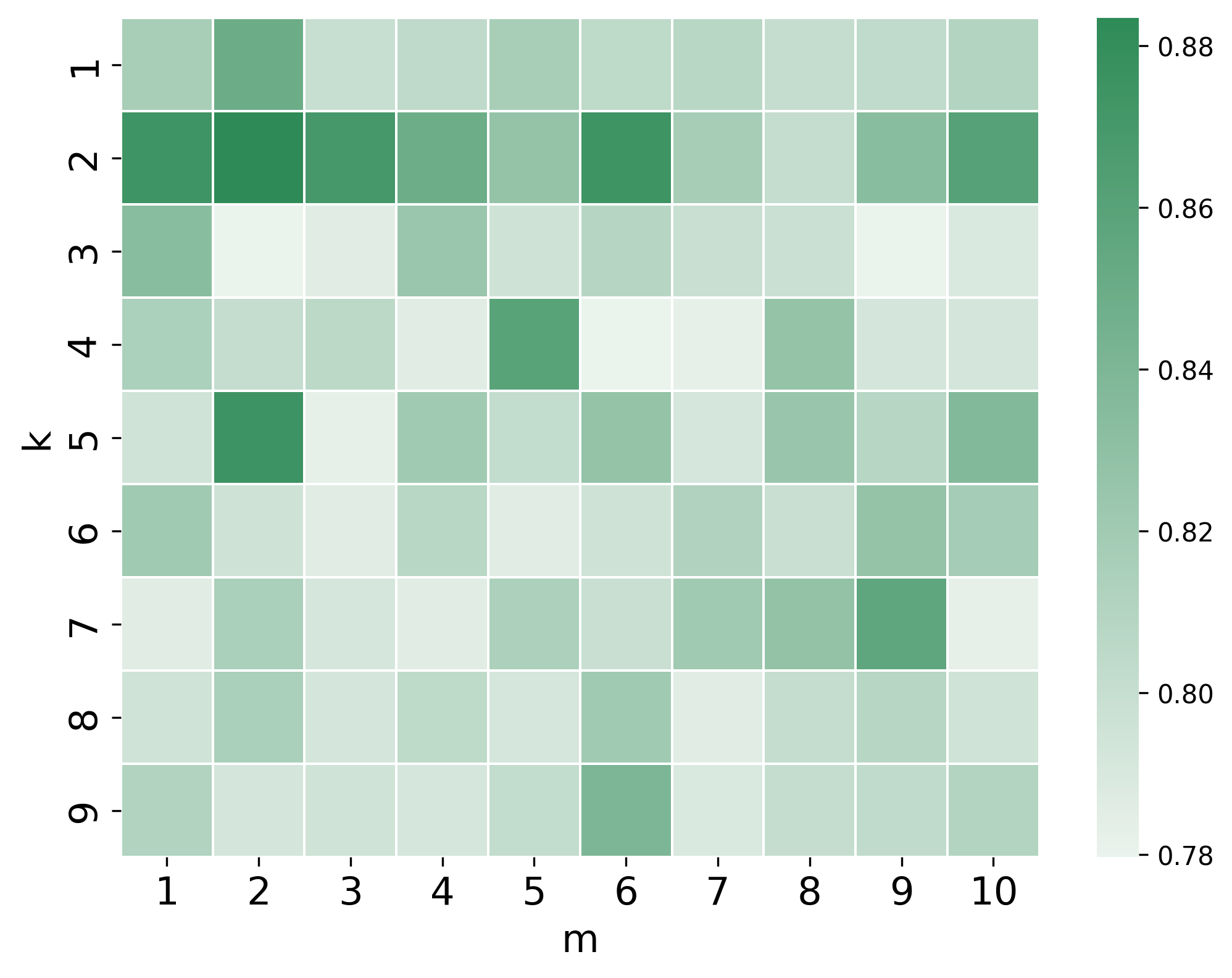}
    \caption{Change in accuracy with varying number of augmentation $k$ and multi-view remasking $m$.}
    \label{abl:multiview_pol}
\end{figure}

\paragraph{Change in results with varying number of augmentations $k$ and multi-view remaskings $m$:} We study the change in classification accuracy with different numbers of augmentations and remaskings for the PolitiFact dataset (Figure \ref{abl:multiview_pol}). We can infer from the figure that the best results are obtained when we set $k = 2$ and $m \leq 6$. This shows that multi-view remaskings help the model achieve superior performance, but more than three remaskings do not bring considerable improvements.

% \begin{table}
%     \centering
% \caption{Accuracy Score for different components of the model}
% \label{tab:my_label}
%     \begin{tabular}{|c|c|c|c|c|c|c|c|c|} \hline 
%          Model&  \multicolumn{4}{|c|}{PolitiFact}& \multicolumn{4}{|c|}{GossipCop}\\ \hline 
%  & Acc& Prec& Rec& F1& Acc& Prec& Rec&F1\\ \hline 
%          Mistral - Zero Shot&  0.747& 0.822& 0.519&0.636&  0.610& 0.731& 0.205& 0.320\\ \hline 
%  LLaMA - Zero Shot& 0.804& 0.826& 0.684& 0.749& 0.680& 0.769& 0.410&0.535\\ \hline 
%  Only $\mathcal{L}_{LGCL} + $ Mistral& 0.822& 0.870& 0.794& 0.830& 0.934& 0.940& 0.924&0.932\\ \hline 
%  Only $\mathcal{L}_{LGCL} + $ LLaMA& 0.841& 0.889& 0.774& 0.828& 0.9488& 0.9560& 0.9422&0.9491\\ \hline 
%          Only $\mathcal{L}_{prop}$&   0.846& 0.849& 0.840&0.845&  0.946& 0.951& 0.939& 0.945\\ \hline 
%  $\mathcal{L}_{LGCL} + \mathcal{L}_{prop} +$ Mistral&  0.873& 0.873& 0.828&0.849& 0.938& 0.941& 0.936& 0.938\\ \hline 
%  $\mathcal{L}_{LGCL} + \mathcal{L}_{prop} + $ LLaMA& \textbf{0.893}& \textbf{0.892}& \textbf{0.897}& \textbf{0.892}& \textbf{0.968} & \textbf{0.965}& \textbf{0.967}&\textbf{0.966}\\ \hline
%     \end{tabular}
% \end{table}
\paragraph{Change in classification results with different components of our model:}
In Table \ref{abl:components}, we show the importance of different components of our model. All the results shown here use $80\%$ labelled data in the final linear SVM for training. As we can see from the table, $\mathcal{L}_{lgcl}$ and $\mathcal{L}_{prop}$ individually produce comparable results. But we get significant improvements in classification accuracy when we combine features generated using $\mathcal{L} = \mathcal{L}_{lgcl}$ + $\mathcal{L}_{prop}$. We also compare the performance of our model with varying versions of the LLM. We use two popular models, Mistral-7B and LLaMA-7B. We show the results when we use the LLMs independently for zero-shot classification. Our model significantly improves the classification results using information from the LLM. One must also note that there is a significant difference between the results from the two LLMs when used independently. But, when used with any component of our model, this difference reduces, thus showing the robustness of the extracted features by the proposed method. 
\begin{table}
    \centering
    \small
\caption{Accuracy Score for different components of the model.}
\label{abl:components}
\resizebox{\linewidth}{!}{%
    \begin{tabular}{c|c|c|c|c} \hline 
        \multirow{2}{*}{Model}&  \multicolumn{2}{|c|}{PolitiFact}& \multicolumn{2}{|c}{GossipCop}\\  
\cline { 2 - 5 } & Acc& F1& Acc&F1\\ \hline 
         Mistral (Zero-shot)&  0.747&0.636&  0.610& 0.320\\ \hline 
 LLaMA (Zero-shot)& 0.804& 0.749& 0.680&0.535\\ \hline 
 Only $\mathcal{L}_{lgcl} + $ Mistral& 0.822& 0.830& 0.934&0.932\\ \hline 
 Only $\mathcal{L}_{lgcl} + $ LLaMA& 0.841& 0.828& 0.948&0.949\\ \hline 
         Only $\mathcal{L}_{prop}$&   0.846&0.845&  0.946& 0.945\\ \hline 
 $\mathcal{L}_{lgcl} + \mathcal{L}_{prop} +$ Mistral&  0.893&0.892& 0.938& 0.938\\ \hline 
 $\mathcal{L}_{lgcl} + \mathcal{L}_{prop} + $ LLaMA& \textbf{0.919}& \textbf{0.918}& \textbf{0.968} &\textbf{0.966}\\ \hline
    \end{tabular}
    }
\end{table}

\begin{figure}[h!]
    \centering
    \includegraphics[width=\linewidth]{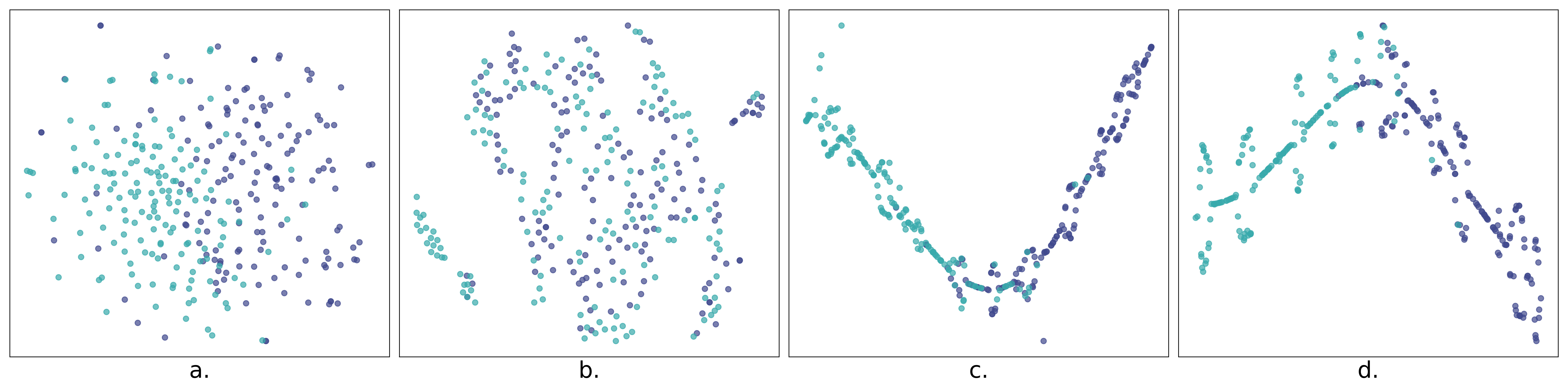}
    \includegraphics[width=\linewidth]{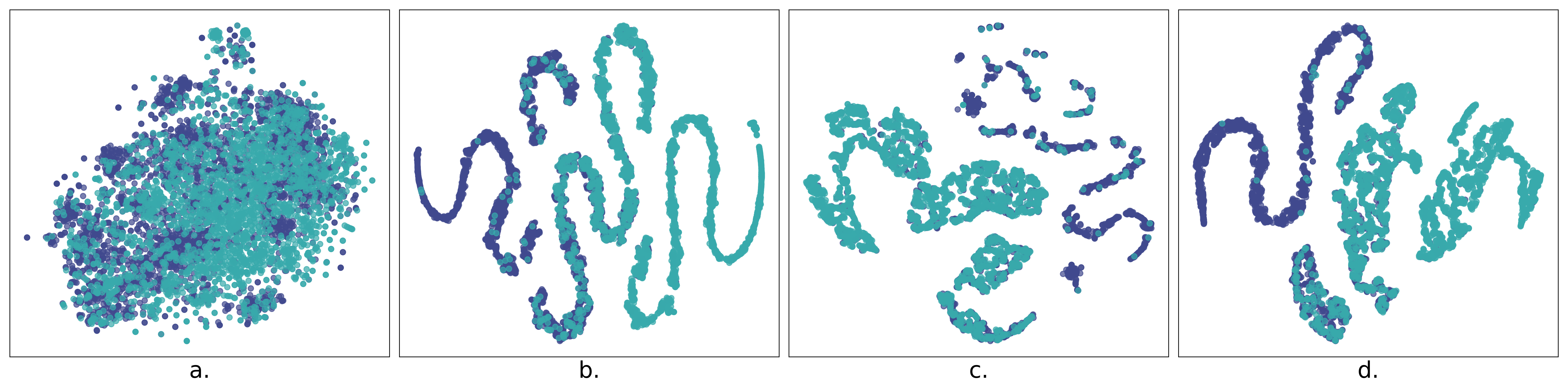}
    \caption{The TSNE plots showing the embeddings of PolitiFact (Row1) and GossipCop (Row2). }
    
    % dataset at various stages. The first column represents the TSNE plot of the original news features, the second column shows the TSNE plot of the original features after an MLP layer, the third column is the TSNE plot of the features generated after the first stage of the pipeline ($\mathcal{H}^{\mathcal{G}^{amr}}$) followed by an MLP layer, the last column represents the TSNE plot of the final features ($\mathcal{H}^{\mathcal{G}^{amr}}$.$\mathcal{H}^{\mathcal{G}^{prop}}$) followed by a linear layer.}
    \label{abl:qualitative_tsne}
\end{figure}

\paragraph{Qualitative results at different stages of our proposed pipeline}
In Figure \ref{abl:qualitative_tsne} we show the feature separation between the real and fake news at different stages of our proposed pipeline. In the first row of the Figure we see the results of PolitiFact dataset and the second row we show the results of the GossipCop dataset. The first column of each row shows the TSNE embedding of the initial features. The second column shows the TSNE plot of the original features after a  single fully connected linear layer (MLP). The third column shows the TSNE plot of the features obtained after the self-supervised AMR graph learning ($\mathcal{H}^{\mathcal{G}^{amr}}$) phase trained with a linear layer. The last columns shows the TSNE plot of the final concatenated features after self-supervised AMR graph learning and multi-view propagation graph learning ($\mathcal{H}^{\mathcal{G}^{amr}}$.$\mathcal{H}^{\mathcal{G}^{prop}}$) with a linear layer. In all the cases we train the MLP with $80\%$ labelled data. 

To quantify the clustering quality, we compute the \textbf{silhouette score} at each stage. For the PolitiFact dataset, the silhouette scores are 0.33, 0.54, 0.62, and 0.64, respectively, indicating progressively better separation between real and fake news as the pipeline advances. Similarly, for the GossipCop dataset, the silhouette scores are 0.16, 0.34, 0.38, and 0.40, again demonstrating consistent improvement. These quantitative results further support the visual evidence, confirming that our model increasingly enhances feature discriminability at each stage of the pipeline.

\section{Conclusion}
This study presents a novel self-supervised approach for misinformation detection. The LLM-based contrastive self-supervised AMR learning framework captures complex semantic relationships in text. This method enhances feature separation between real and fake news by leveraging an LLM-based negative sampler. Additionally, we introduce a multi-view graph-masked autoencoder that integrates social context and news propagation patterns for more robust detection. Through extensive experiments, the proposed method is found to produce state-of-the-art performance. Beyond misinformation detection, our methodology has broader applications in NLP. For instance, self-supervised AMR graph learning can be applied to tasks like question-answering and event detection, while multi-view social context and propagation graph learning can be leveraged for hate speech and aggression detection, etc. This work not only advances misinformation detection but also lays the groundwork for tackling various NLP challenges using graph-based learning in constraint settings.

\bibliography{aaai2026}

\begin{thebibliography}{40}
\providecommand{\natexlab}[1]{#1}

\bibitem[{Banarescu et~al.(2013)Banarescu, Bonial, Cai, Georgescu, Griffitt, Hermjakob, Knight, Koehn, Palmer, and Schneider}]{banarescu2013}
Banarescu, L.; Bonial, C.; Cai, S.; Georgescu, M.; Griffitt, K.; Hermjakob, U.; Knight, K.; Koehn, P.; Palmer, M.; and Schneider, N. 2013.
\newblock {A}bstract {M}eaning {R}epresentation for Sembanking.
\newblock In \emph{Proceedings of the 7th Linguistic Annotation Workshop and Interoperability with Discourse}, 178--186. Sofia, Bulgaria.

\bibitem[{Cai and Lam(2020)}]{cai2020}
Cai, D.; and Lam, W. 2020.
\newblock Graph Transformer for Graph-to-Sequence Learning.
\newblock In \emph{AAAI}, 7464--7471. {AAAI} Press.

\bibitem[{Cho et~al.(2014)Cho, van Merri{\"e}nboer, Gulcehre, Bahdanau, Bougares, Schwenk, and Bengio}]{cho-etal-2014-learning}
Cho, K.; van Merri{\"e}nboer, B.; Gulcehre, C.; Bahdanau, D.; Bougares, F.; Schwenk, H.; and Bengio, Y. 2014.
\newblock Learning Phrase Representations using {RNN} Encoder{--}Decoder for Statistical Machine Translation.
\newblock In \emph{EMNLP}, 1724--1734. Doha, Qatar: ACL.

\bibitem[{Devlin et~al.(2019)Devlin, Chang, Lee, and Toutanova}]{devlin-etal-2019-bert}
Devlin, J.; Chang, M.-W.; Lee, K.; and Toutanova, K. 2019.
\newblock {BERT}: Pre-training of Deep Bidirectional Transformers for Language Understanding.
\newblock In Burstein, J.; Doran, C.; and Solorio, T., eds., \emph{Proceedings of the 2019 Conference of the North {A}merican Chapter of the Association for Computational Linguistics: Human Language Technologies, Volume 1 (Long and Short Papers)}, 4171--4186. Minneapolis, Minnesota: Association for Computational Linguistics.

\bibitem[{Dewatana and Adillah(2021)}]{dewatana2021effectiveness}
Dewatana, H.; and Adillah, S.~U. 2021.
\newblock The effectiveness of criminal eradication on hoax information and fake news.
\newblock \emph{Law Development Journal}, 3(3): 513--520.

\bibitem[{Dou et~al.(2021)Dou, Shu, Xia, Yu, and Sun}]{10.1145/3404835.3462990}
Dou, Y.; Shu, K.; Xia, C.; Yu, P.~S.; and Sun, L. 2021.
\newblock User Preference-aware Fake News Detection.
\newblock In \emph{Proceedings of the 44th International ACM SIGIR Conference on Research and Development in Information Retrieval}, SIGIR '21, 2051–2055. New York, NY, USA: Association for Computing Machinery.
\newblock ISBN 9781450380379.

\bibitem[{Dun et~al.(2021)Dun, Tu, Chen, Hou, and Yuan}]{dun2021}
Dun, Y.; Tu, K.; Chen, C.; Hou, C.; and Yuan, X. 2021.
\newblock KAN: Knowledge-aware Attention Network for Fake News Detection.
\newblock \emph{AAAI}, 35(1): 81--89.

\bibitem[{Feng, Banerjee, and Choi(2012)}]{feng-etal-2012-syntactic}
Feng, S.; Banerjee, R.; and Choi, Y. 2012.
\newblock Syntactic Stylometry for Deception Detection.
\newblock In \emph{ACL (Volume 2: Short Papers)}, 171--175. Jeju Island, Korea: ACL.

\bibitem[{Gangireddy et~al.(2020)Gangireddy, P, Long, and Chakraborty}]{gang2020}
Gangireddy, S. C.~R.; P, D.; Long, C.; and Chakraborty, T. 2020.
\newblock Unsupervised Fake News Detection: A Graph-based Approach.
\newblock In \emph{Proceedings of the 31st ACM Conference on Hypertext and Social Media}, HT '20, 75–83. New York, NY, USA: Association for Computing Machinery.
\newblock ISBN 9781450370981.

\bibitem[{Ghanem et~al.(2021)Ghanem, Ponzetto, Rosso, and Rangel}]{ghanem2021fakeflow}
Ghanem, B.; Ponzetto, S.~P.; Rosso, P.; and Rangel, F. 2021.
\newblock FakeFlow: Fake News Detection by Modeling the Flow of Affective Information.
\newblock In \emph{16th EACL}.

\bibitem[{Gupta, Rajora, and Kundu(2025)}]{ea2n2025}
Gupta, S.; Rajora, A.; and Kundu, S. 2025.
\newblock EA2N: Evidence-based AMR Attention Network for Fake News Detection.
\newblock \emph{IEEE Transactions on Knowledge and Data Engineering}, 1--12.

\bibitem[{Gupta et~al.(2023)Gupta, Yadav, Kundu, and Sankepally}]{Gupta2023}
Gupta, S.; Yadav, N.; Kundu, S.; and Sankepally, S. 2023.
\newblock FakEDAMR: Fake News Detection Using Abstract Meaning Representation Network.
\newblock In \emph{International Conference on Complex Networks and Their Applications}, 308--319. Springer.

\bibitem[{Hou et~al.(2023)Hou, He, Cen, Liu, Dong, Kharlamov, and Tang}]{10.1145/3543507.3583379}
Hou, Z.; He, Y.; Cen, Y.; Liu, X.; Dong, Y.; Kharlamov, E.; and Tang, J. 2023.
\newblock GraphMAE2: A Decoding-Enhanced Masked Self-Supervised Graph Learner.
\newblock In \emph{Proceedings of the ACM Web Conference 2023}, WWW '23, 737–746. New York, NY, USA: Association for Computing Machinery.
\newblock ISBN 9781450394161.

\bibitem[{Hou et~al.(2022)Hou, Liu, Cen, Dong, Yang, Wang, and Tang}]{10.1145/3534678.3539321}
Hou, Z.; Liu, X.; Cen, Y.; Dong, Y.; Yang, H.; Wang, C.; and Tang, J. 2022.
\newblock GraphMAE: Self-Supervised Masked Graph Autoencoders.
\newblock In \emph{Proceedings of the 28th ACM SIGKDD Conference on Knowledge Discovery and Data Mining}, KDD '22, 594–604. New York, NY, USA: Association for Computing Machinery.
\newblock ISBN 9781450393850.

\bibitem[{Hu et~al.(2021)Hu, Yang, Zhang, Zhong, Tang, Shi, Duan, and Zhou}]{hu-etal-2021-compare}
Hu, L.; Yang, T.; Zhang, L.; Zhong, W.; Tang, D.; Shi, C.; Duan, N.; and Zhou, M. 2021.
\newblock Compare to The Knowledge: Graph Neural Fake News Detection with External Knowledge.
\newblock In \emph{ACL-IJCNLP (Volume 1: Long Papers)}, 754--763. Online: ACL.

\bibitem[{Kipf and Welling(2016)}]{kipf2016variationalgraphautoencoders}
Kipf, T.~N.; and Welling, M. 2016.
\newblock Variational Graph Auto-Encoders.
\newblock arXiv:1611.07308.

\bibitem[{Li et~al.(2021)Li, Guo, Wang, and Zheng}]{li2021}
Li, D.; Guo, H.; Wang, Z.; and Zheng, Z. 2021.
\newblock Unsupervised Fake News Detection Based on Autoencoder.
\newblock \emph{IEEE Access}, 9: 29356--29365.

\bibitem[{Li et~al.(2024)Li, Li, Luvembe, and Tong}]{li2024}
Li, S.; Li, W.; Luvembe, A.~M.; and Tong, W. 2024.
\newblock Graph Contrastive Learning With Feature Augmentation for Rumor Detection.
\newblock \emph{IEEE Transactions on Computational Social Systems}, 11(4): 5158--5167.

\bibitem[{Li et~al.(2022)Li, Liu, Yang, Peng, and Zhou}]{li2022}
Li, Z.; Liu, F.; Yang, W.; Peng, S.; and Zhou, J. 2022.
\newblock A Survey of Convolutional Neural Networks: Analysis, Applications, and Prospects.
\newblock \emph{IEEE Transactions on Neural Networks and Learning Systems}, 33(12): 6999--7019.

\bibitem[{Liu and Wu(2018)}]{Liu_Wu_2018}
Liu, Y.; and Wu, Y.-F. 2018.
\newblock Early Detection of Fake News on Social Media Through Propagation Path Classification with Recurrent and Convolutional Networks.
\newblock \emph{AAAI}, 32(1).

\bibitem[{Long et~al.(2017)Long, Lu, Xiang, Li, and Huang}]{long-etal-2017-fake}
Long, Y.; Lu, Q.; Xiang, R.; Li, M.; and Huang, C.-R. 2017.
\newblock Fake News Detection Through Multi-Perspective Speaker Profiles.
\newblock In \emph{IJCNLP (Volume 2: Short Papers)}, 252--256. Taipei, Taiwan: Asian Federation of Natural Language Processing.

\bibitem[{Ma et~al.(2016{\natexlab{a}})Ma, Gao, Mitra, Kwon, Jansen, Wong, and Cha}]{ma2016}
Ma, J.; Gao, W.; Mitra, P.; Kwon, S.; Jansen, B.~J.; Wong, K.-F.; and Cha, M. 2016{\natexlab{a}}.
\newblock Detecting rumors from microblogs with recurrent neural networks.
\newblock In \emph{Proceedings of the Twenty-Fifth International Joint Conference on Artificial Intelligence}, IJCAI'16, 3818–3824. AAAI Press.
\newblock ISBN 9781577357704.

\bibitem[{Ma et~al.(2016{\natexlab{b}})Ma, Gao, Mitra, Kwon, Jansen, Wong, and Cha}]{jing2016}
Ma, J.; Gao, W.; Mitra, P.; Kwon, S.; Jansen, B.~J.; Wong, K.-F.; and Cha, M. 2016{\natexlab{b}}.
\newblock Detecting Rumors from Microblogs with Recurrent Neural Networks.
\newblock In \emph{IJCAI}, IJCAI'16, 3818–3824. AAAI Press.
\newblock ISBN 9781577357704.

\bibitem[{Min et~al.(2022)Min, Rong, Bian, Xu, Zhao, Huang, and Ananiadou}]{min2022}
Min, E.; Rong, Y.; Bian, Y.; Xu, T.; Zhao, P.; Huang, J.; and Ananiadou, S. 2022.
\newblock Divide-and-Conquer: Post-User Interaction Network for Fake News Detection on Social Media.
\newblock In \emph{Proceedings of the ACM Web Conference 2022}, WWW '22, 1148–1158. New York, NY, USA: Association for Computing Machinery.
\newblock ISBN 9781450390965.

\bibitem[{Popat et~al.(2017)Popat, Mukherjee, Str\"{o}tgen, and Weikum}]{popat2017}
Popat, K.; Mukherjee, S.; Str\"{o}tgen, J.; and Weikum, G. 2017.
\newblock Where the Truth Lies: Explaining the Credibility of Emerging Claims on the Web and Social Media.
\newblock WWW '17 Companion, 1003–1012. Republic and Canton of Geneva, CHE: International World Wide Web Conferences Steering Committee.
\newblock ISBN 9781450349147.

\bibitem[{Shu et~al.(2019)Shu, Cui, Wang, Lee, and Liu}]{shu2019}
Shu, K.; Cui, L.; Wang, S.; Lee, D.; and Liu, H. 2019.
\newblock dEFEND: Explainable Fake News Detection.
\newblock In \emph{Proceedings of the 25th ACM SIGKDD International Conference on Knowledge Discovery \& Data Mining}, KDD '19, 395–405. New York, NY, USA: Association for Computing Machinery.
\newblock ISBN 9781450362016.

\bibitem[{Shu et~al.(2020)Shu, Mahudeswaran, Wang, Lee, and Liu}]{kai2020}
Shu, K.; Mahudeswaran, D.; Wang, S.; Lee, D.; and Liu, H. 2020.
\newblock FakeNewsNet: A Data Repository with News Content, Social Context, and Spatiotemporal Information for Studying Fake News on Social Media.
\newblock \emph{Big Data}, 8(3): 171--188.

\bibitem[{Silva et~al.(2024)Silva, Luo, Karunasekera, and Leckie}]{silva2024}
Silva, A.; Luo, L.; Karunasekera, S.; and Leckie, C. 2024.
\newblock { Unsupervised Domain-Agnostic Fake News Detection Using Multi-Modal Weak Signals }.
\newblock \emph{IEEE Transactions on Knowledge \& Data Engineering}, 36(11): 7283--7295.

\bibitem[{Sun et~al.(2022)Sun, Qian, Dong, Li, and Zhu}]{sun2022}
Sun, T.; Qian, Z.; Dong, S.; Li, P.; and Zhu, Q. 2022.
\newblock Rumor Detection on Social Media with Graph Adversarial Contrastive Learning.
\newblock In \emph{Proceedings of the ACM Web Conference 2022}, WWW '22, 2789–2797. New York, NY, USA: Association for Computing Machinery.
\newblock ISBN 9781450390965.

\bibitem[{Wang et~al.(2018)Wang, Ma, Jin, Yuan, Xun, Jha, Su, and Gao}]{wang2018}
Wang, Y.; Ma, F.; Jin, Z.; Yuan, Y.; Xun, G.; Jha, K.; Su, L.; and Gao, J. 2018.
\newblock EANN: Event Adversarial Neural Networks for Multi-Modal Fake News Detection.
\newblock In \emph{Proceedings of the 24th ACM SIGKDD International Conference on Knowledge Discovery \& Data Mining}, KDD '18, 849–857. New York, NY, USA: Association for Computing Machinery.
\newblock ISBN 9781450355520.

\bibitem[{Wu et~al.(2023)Wu, Lin, Tan, Gao, and Li}]{wu2021}
Wu, L.; Lin, H.; Tan, C.; Gao, Z.; and Li, S.~Z. 2023.
\newblock Self-Supervised Learning on Graphs: Contrastive, Generative, or Predictive.
\newblock 35(4): 4216–4235.

\bibitem[{Xu et~al.(2019)Xu, Hu, Leskovec, and Jegelka}]{xu2018how}
Xu, K.; Hu, W.; Leskovec, J.; and Jegelka, S. 2019.
\newblock How Powerful are Graph Neural Networks?
\newblock In \emph{International Conference on Learning Representations}.

\bibitem[{Yang et~al.(2022)Yang, Wang, Jin, Li, Lian, and Xie}]{yang2022}
Yang, R.; Wang, X.; Jin, Y.; Li, C.; Lian, J.; and Xie, X. 2022.
\newblock Reinforcement Subgraph Reasoning for Fake News Detection.
\newblock In \emph{Proceedings of the 28th ACM SIGKDD Conference on Knowledge Discovery and Data Mining}, KDD '22, 2253–2262. New York, NY, USA: Association for Computing Machinery.
\newblock ISBN 9781450393850.

\bibitem[{Yao et~al.(2021)Yao, Wu, Al-Shedivat, and Xing}]{yao2021}
Yao, H.; Wu, Y.-x.; Al-Shedivat, M.; and Xing, E. 2021.
\newblock Knowledge-Aware Meta-learning for Low-Resource Text Classification.
\newblock In \emph{Proceedings of the 2021 Conference on Empirical Methods in Natural Language Processing}, 1814--1821. Online and Punta Cana, Dominican Republic: Association for Computational Linguistics.

\bibitem[{Yin et~al.(2024)Yin, Zhu, Wu, Gao, and Wang}]{Yin_Zhu_Wu_Gao_Wang_2024}
Yin, S.; Zhu, P.; Wu, L.; Gao, C.; and Wang, Z. 2024.
\newblock GAMC: An Unsupervised Method for Fake News Detection Using Graph Autoencoder with Masking.
\newblock \emph{Proceedings of the AAAI Conference on Artificial Intelligence}, 38(1): 347--355.

\bibitem[{Yin, Han, and Yu(2008)}]{yin2008}
Yin, X.; Han, J.; and Yu, P.~S. 2008.
\newblock Truth Discovery with Multiple Conflicting Information Providers on the Web.
\newblock \emph{IEEE Transactions on Knowledge and Data Engineering}, 20(6): 796--808.

\bibitem[{Yuan et~al.(2019)Yuan, Ma, Zhou, Han, and Hu}]{yuan2019}
Yuan, C.; Ma, Q.; Zhou, W.; Han, J.; and Hu, S. 2019.
\newblock { Jointly Embedding the Local and Global Relations of Heterogeneous Graph for Rumor Detection }.
\newblock In \emph{2019 IEEE International Conference on Data Mining (ICDM)}, 796--805. Los Alamitos, CA, USA: IEEE Computer Society.

\bibitem[{Zhang et~al.(2019)Zhang, Ma, Duh, and Van~Durme}]{zhang-etal-2019-amr}
Zhang, S.; Ma, X.; Duh, K.; and Van~Durme, B. 2019.
\newblock {AMR} Parsing as Sequence-to-Graph Transduction.
\newblock In \emph{ACL}, 80--94. Florence, Italy: ACL.

\bibitem[{Zhang et~al.(2023)Zhang, Trinh, Cao, Cui, and Liu}]{zhang2023detecting}
Zhang, Y.; Trinh, L.; Cao, D.; Cui, Z.; and Liu, Y. 2023.
\newblock Detecting Out-of-Context Multimodal Misinformation with interpretable neural-symbolic model.
\newblock arXiv:2304.07633.

\bibitem[{Zhou, Wu, and Zafarani(2020)}]{zhou2020}
Zhou, X.; Wu, J.; and Zafarani, R. 2020.
\newblock SAFE: Similarity-Aware Multi-modal Fake News Detection.
\newblock In Lauw, H.~W.; Wong, R. C.-W.; Ntoulas, A.; Lim, E.-P.; Ng, S.-K.; and Pan, S.~J., eds., \emph{Advances in Knowledge Discovery and Data Mining}, 354--367. Cham: Springer International Publishing.
\newblock ISBN 978-3-030-47436-2.

\end{thebibliography}
\appendix

\section{Details on Datasets and Implementation}
PolitiFact is dedicated to news coverage revolving around U.S. political affairs, while GossipCop delves into stories about Hollywood celebrities. These datasets also capture the broader social dynamics by including information about how news spreads through networks and the posting patterns of users. We evaluate our model using a set of metrics, including Precision (Pre), Recall (Rec), F1-score, and Accuracy (Acc). Comprehensive details of the datasets are provided in Table \ref{tab1}.
\begin{table}[ht]
\centering
\caption{Datasets Statistics}
\resizebox{\linewidth}{!}{%
\begin{tabular}{c|c|c|c|c|c}
\hline & \# News & \# True & \# Fake & \# Nodes & \# Edges \\
\hline PolitiFact & 314 & 157 & 157 &  41054 & 40740\\
GossipCop & 5464 & 2732 & 2732 &  314262 & 308798\\
\hline
\end{tabular}
}
\label{tab1}
\end{table}

\paragraph{Implementation Details: }
In order to generate the AMR graph, we have used a pretrained STOG model \cite{zhang-etal-2019-amr}. For LGCL, we use $\alpha=0.5$  and in order to integrate the evidence in the AMR graph, we use the same parameters described in \cite{ea2n2025}. For social context and propagation graph learning we use $2$ encoder layers and $1$ decoder layer. For multi-view remasking, we select $k=2$ and $m=2$. We selected Support Vector Machine (SVM) as the final classifier and reported the results from 80 \% of the training data with 5-fold cross-validation. Although we provided our results for each test size percentage in result table, our main results are based on an 80:20 train-test split to ensure consistency with other methods. We have trained our model on RTX A5000 Nvidia GPU with 24 GB GPU memory. The training of AMR took 1 hour for PolitiFact and took 3 hours for the GossipCop dataset with 50 epochs. Multi-view masked graph learning took 5 mins for the PolitiFact dataset and 15 minutes for the GossipCop dataset.


\end{document}